\documentclass[sigconf, screen]{acmart}
\renewcommand\footnotetextcopyrightpermission[1]{}
\usepackage{hyperref}
\usepackage{enumitem}
\usepackage{microtype}
\usepackage{graphicx}
\usepackage{subcaption}
\usepackage{multirow}
\usepackage{booktabs} 
\usepackage{array}
\usepackage[table]{xcolor}
\usepackage{amsmath}
\usepackage{tabularx}
\usepackage{mathtools}
\usepackage{amsthm}
\usepackage{fontawesome5}
\usepackage{bbding}
\AtBeginDocument{%
  }

\begin{document}

\title{Enhancing Foundation VLM Robustness to Missing Modality: Scalable Diffusion for Bi-directional Feature Restoration}


\settopmatter{authorsperrow=4}
\author{Wei Dai}
\affiliation{%
  \institution{Department of Electronics and Information \\Xi'an Jiaotong University}
  \city{Xi'an}
  \country{China}}
\email{daiwei946@gmail.com}

\author{Haoyu Wang}
\affiliation{%
  \institution{Department of Electronics and Information \\Xi'an Jiaotong University}
  \city{Xi'an}
  \country{China}}

\author{Honghao Chang}
\affiliation{%
  \institution{School of Information and Communications Engineering \\Xi'an Jiaotong University}
  \city{Xi'an}
  \country{China}}

\author{Lijun He}
\affiliation{%
  \institution{School of Information and Communications Engineering \\Xi'an Jiaotong University}
  \city{Xi'an}
  \country{China}}

\author{Fan Li}
\affiliation{%
  \institution{School of Information and Communications Engineering \\Xi'an Jiaotong University}
  \city{Xi'an}
  \country{China}}

\author{Jian Sun}
\affiliation{%
  \institution{School of
Mathematics and Statistics \\Xi'an Jiaotong University}
  \city{Xi'an}
  \country{China}}

\author{Haixia Bi$^{\dagger}$}
\thanks{\small $^{\dagger}$Corresponding Author. \hspace{0.15mm} \faEnvelope[regular]~ haixia.bi@xjtu.edu.cn}
\affiliation{%
  \institution{School of Information and Communications Engineering \\Xi'an Jiaotong University}
  \city{Xi'an}
  \country{China}}


\renewcommand{\shortauthors}{}

\begin{abstract}
 Vision Language Model (VLM) typically assume complete modality input during inference. However, their effectiveness drops sharply when certain modalities are unavailable or incomplete. Current research on missing modality primarily faces two dilemmas: Prompt-based methods struggle to restore missing yet indispensable features and degrade the generalizability of VLM. Imputation-based approaches, lacking effective guidance, are prone to generating semantically irrelevant noise. Restoring precise semantics while sustaining VLM’s generalization remains challenging. Therefore, we propose a general missing modality restoration strategy in this paper. We introduce an enhanced diffusion model as a pluggable mid-stage training module to effectively restore missing features. Our strategy introduces two key innovations: (I) Dynamic Modality Gating, which adaptively leverages conditional features to guide the generation of semantically consistent features; (II) Cross-Modal Mutual Learning mechanism, which bridges the semantic spaces of the dual models to achieve bi-directional alignment. Notably, our strategy maintains the original integrity of the pre-trained VLM, requiring no fine-tuning of the backbone models while significantly boosting resilience to information loss. Zero-shot evaluations across benchmark datasets demonstrate that our approach consistently outperforms existing baselines, establishing it as a robust and scalable extension that ensures VLM reliability across diverse missing rates and conditions. Our code and models will be publicly available.
\end{abstract}






\maketitle

\section{Introduction}

Multimodal learning~\cite{baltruvsaitis2018multimodal} has achieved remarkable success~\cite{ramesh2022hierarchical,liu2023visual} in the fields of vision and language~\cite{li2022grounded,alayrac2022flamingo}. In particular, foundational Vision Language Models (VLMs) represented by CLIP~\cite{radford2021learning} and BLIP~\cite{li2022blip} have constructed a cross-modal semantic space through large-scale contrastive pretraining. However, these models are typically built on the idealized assumption that all modalities are fully available during the inference phase. In real-world deployment, missing modality has become a critical bottleneck constraining the practical performance of VLMs due to factors such as collection costs and privacy protection. 

\begin{figure}[t]
\vspace{10pt}
    \centering
    \includegraphics[width=0.5\textwidth, trim=1cm 0cm 1cm 0cm, clip]{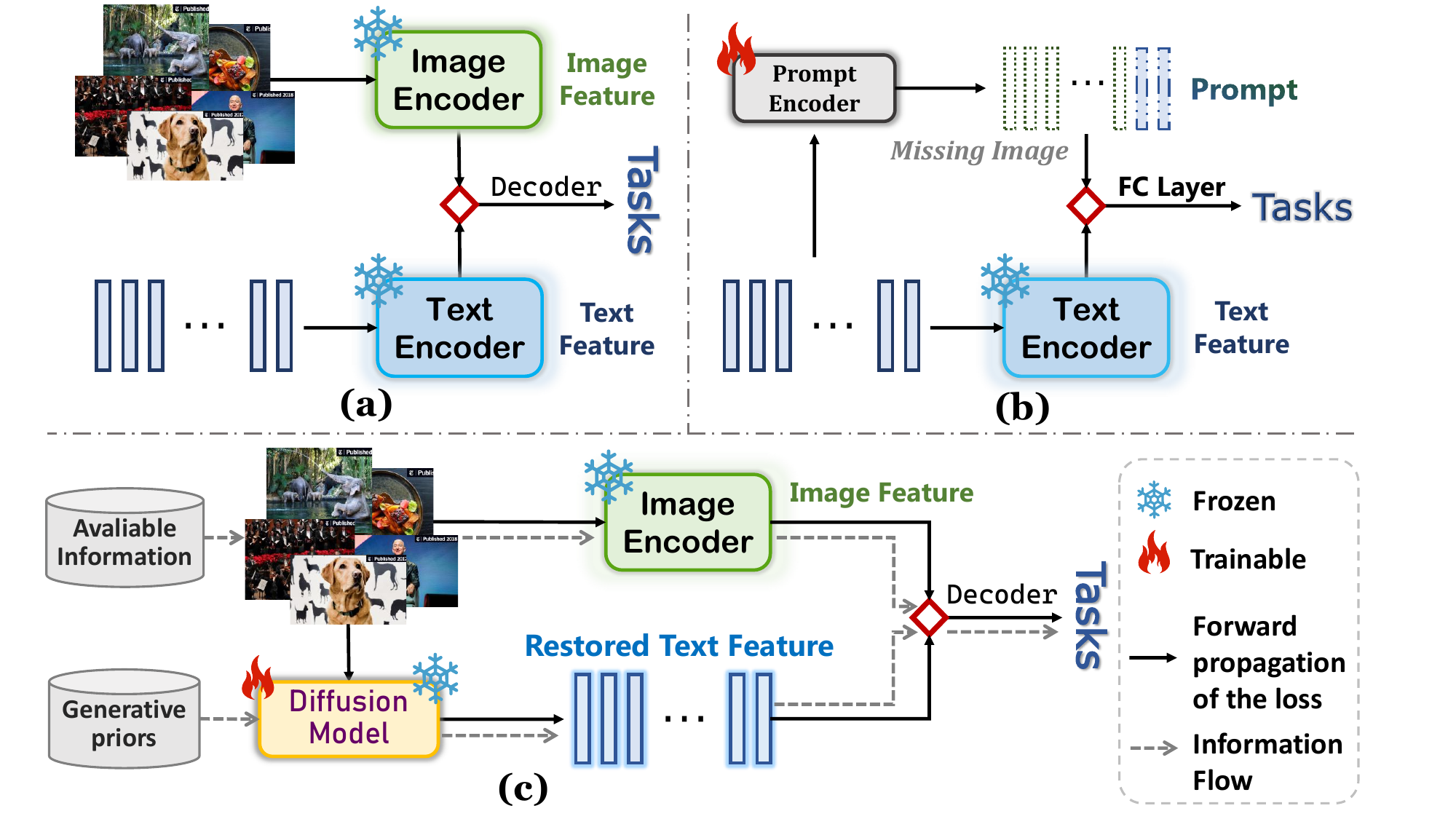}
\vspace{-15pt}
    \caption{An overview of Multimodal Missing Modalitiy: (a) Foundation VLM working with complete modality; (b) Prompt learning methods for missing modality; (c) (Ours) Diffusion feature restoration method for missing modality.}
    \Description{An overview of Multimodal Missing Modalitiy}
    \label{fig:subfigure}
  \vspace{-5pt}
\end{figure}

Existing research on missing modalities primarily falls into two categories: Prompt-based methods~\cite{shi2024deep,zhang2025synergistic,zhao2025enhancing} adapt to missing modalities via learnable prompts~\cite{jang2024towards,lee2023multimodal,jia2022visual}, but often compromise the VLM generalization by introducing task-specific parameters and altering the model's original architectures. As a result, these methods provide only dataset-dependent adaptations rather than true feature restoration, limiting their utility in tasks requiring complete features. Imputation-based methods~\cite{ma2021smil,xu2025mcmoe,kim2024missing} attempt to reconstruct the missing modalities~\cite{lang2025retrieval,lang2025redeeming}. But prior generative models are prone to producing semantically irrelevant noise in the absence of strong guidance. A core challenge lies in restoring missing features accurately while maintaining consistency with existing modalities on the deep semantic manifold.

\begin{figure}[t]
    \centering
    \includegraphics[width=0.47\textwidth, trim=0.5cm 0.5cm 0.5cm 0cm, clip]{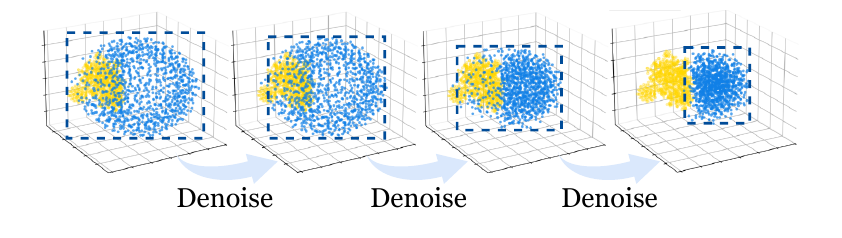}
    \vspace{-7pt}
    \caption{PCA results across 1,000 samples show that our denoising process ($T=1000, 800, 500, 50$) effectively aligns restored text features with ground-truth image embeddings.}
    \Description{PCA results across 1,000 samples}
    \label{fig:diffusion_sample}
    \vspace{-23pt}
\end{figure}

To overcome these limitations, this paper proposes a feature restoration strategy based on a Scalable Diffusion Model. Unlike conventional strategies that rely on shallow MLP-based projections or static constant padding (e.g., filling missing entries with zeros), the diffusion model exploits its robust generative priors to capture the intricate manifold geometry inherent in the VLM's semantic space (Figure~\ref{fig:subfigure}). Through step-by-step denoising (Figure~\ref{fig:diffusion_sample}), the model can completely restore missing VLM's features, making it applicable across various tasks and scenarios, thereby genuinely enhancing the robustness and generalization of VLM under incomplete inputs. By performing mid-stage training within the VLM feature space, we transform the generative prior of the diffusion model into semantic information increment. This increment not only compensates for the lost semantics but also serves as a powerful information augmentation, effectively bridging the semantic gap between unimodal inputs and multimodal representations.

To enable the diffusion model to accurately extract key semantic information while suppressing modality-irrelevant noise, we propose \textbf{Dynamic Modality Gating} mechanism. This mechanism leverages conditional features from available modalities to keep task-relevant information and filter out noise. By utilizing diffusion priors, it enhances the signal-to-noise ratio and maintains semantic consistency throughout the generative process.
To achieve high-quality bidirectional feature restoration under conditions of missing modalities, we propose \textbf{Cross-modal Mutual Learning} mechanism. This mechanism bridges the latent semantic spaces of visual and text encoders to achieve alignment across bidirectional semantic manifolds. It prevents the synthesized features from deviating from the target semantic space, addresses the uncertainty inherent in unidirectional mapping, and reinforces the robustness of both Image-to-Text (I2T) and Text-to-Image (T2I) restoration.

We conducted zero-shot feature restoration experiments on four visison-language benchmark datasets, and the results demonstrate that our restorative mid-stage-trained model outperforms existing strong baselines. This indicates its capability to successfully restore semantically consistent features on unseen datasets, significantly boosting downstream task performance. Furthermore, our strategy demonstrates observable scalability. In summury, our contributions are as follows:

\begin{itemize}

    \item We propose a scalable framework based on diffusion models that moves beyond simplistic feature imputation to achieve high-fidelity semantic restoration within VLM's latent space.

    \item We introduce the Dynamic Modality Gating and Cross-modal Mutual Learning mechanisms, which effectively filter modality noise and enforce bi-directional semantic alignment \\through cyclic consistency.

    \item Through mid-stage training on large datasets, we demonstrate that our approach greatly enhances VLM robustness in incomplete modality scenarios, establishing a new benchmark for building universal missing-robust models.
\end{itemize}

\section{Related Work}

\subsection{Multimodal Missing Modalities}
The problem of missing modality~\cite{zhao2021missing,ma2022multimodal,zhang2025admc,pipoli2025missrag} has emerged as a significant research field in multimodal learning~\cite{li2023blip,yang2025vision,yang2025small,an2025unictokens,an2026genius,lin2024draw,lin2025perceive,zheng2026pearl}. Existing methods can be categorized into two types:

Prompt-based Methods aim to adapt multimodal models to incomplete inputs by introducing a limited number of learnable prompts~\cite{jang2024towards}. Notable examples include missing-aware prompt~\cite{lee2023multimodal}, modality-specific prompts~\cite{jia2022visual}, Deep Correlated Prompting~\cite{shi2024deep}, Synergistic Prompt~\cite{zhang2025synergistic}, and Memory-Driven~Prompts~\cite{zhao2025enhancing}. Besides, PROMISE~\cite{chen2025promise} developed prompt-attention with modality-specific prompt pools. Reflecting the earlier statement, the lack of robust generalizability remains a major drawback of prompt methods.

Generation and Imputation Methods attempt to explicitly reconstruct missing information. Early research, such as SMIL~\cite{ma2021smil}, utilizes Bayesian Meta-learning to handle missing data. Missing Modality Prediction~\cite{kim2024missing} employs trainable prompts to predict missing features, while McMoE~\cite{xu2025mcmoe} utilizes an adaptive gated modality generator network to synthesize missing characteristics. In particular, Retrieval-augmented Methods leverage external knowledge. RAGPT~\cite{lang2025retrieval} framework identifies similar instances through a retriever, and REDEEM~\cite{lang2025redeeming} further introduces retrieval-guided generation utilizing a memory bank. However, previous generation and imputation strategies often face a trade-off between sub-optimal performance and high memory consumption (e.g., memory banks).

\subsection{Diffusion Models for Latent Modeling}

Diffusion Models~\cite{sohl2015deep,peebles2023scalable} have demonstrated great potential in image synthesis~\cite{dhariwal2021diffusion,rombach2022high}, video generation~\cite{ho2022video}, and various discriminative tasks~\cite{li2023your}. Diffusion models gradually transform data into Gaussian noise via a forward diffusion process, subsequently learning a reverse denoising process to recover the original data distribution from the noise~\cite{ho2020denoising}. Compared to GANs~\cite{goodfellow2014generative} and VAEs~\cite{kingma2013auto}, diffusion models can better capture the structures of high-dimensional manifolds, avoiding training instability and model collapse.

The semantic consistency inherent in latent representations generated by diffusion models~\cite{lee2025diffusion,ramesh2022hierarchical,becker2025controlling} has opened new avenues for feature-level augmentation. 
By leveraging these advanced variants, existing studies~\cite{chen2023diffusiondet,amit2021segdiff,yu2024universal} have successfully synthesized features for tasks such as cross-modal retrieval and data augmentation. While prior studies~\cite{wang2023incomplete,dai2025unbiased,kebaili2025amm,meng2024multi} have explored diffusion-based modality imputation in task-specific domains like multimodal sentiment analysis or medical image synthesis, these methods typically rely on multimodal redundancy for fixed classifiers and fail to scale effectively to larger, more complex models. 
Extending such paradigms to foundation VLMs is non-trivial, as the latter require rigorous semantic alignment to maintain zero-shot capabilities. 
Our work transcends these task-oriented constraints by introducing a scalable restoration module. 
As a universal plug-in, it enhances VLM resilience without compromising the original architectural integrity.

\begin{figure*}[t]
    \centering
    \includegraphics[width=1\textwidth, trim=0.7cm 12.9cm 3.6cm 0.45cm, clip]{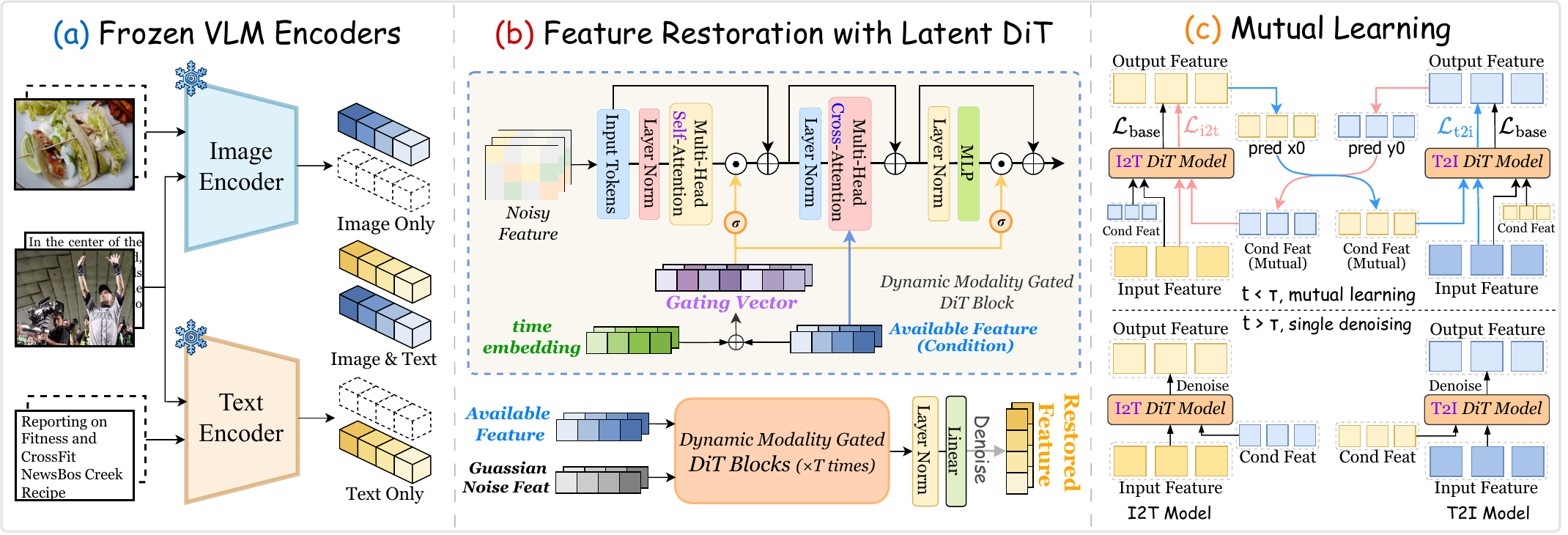}
    \caption{The architecture of our Missing Modality Restoration Framework. (a) Feature extraction process for available modalities utilizing frozen pre-trained VLM Encoders. (b) The top section depicts the gating mechanism of the DiT backbone, and the bottom section depicts the restoration of missing features from noise. (c) Architectural design of the mutual learning mechanism, incorporating multi-objective loss functions within specific denoising intervals.}
    \Description{The architecture of our Missing Modality Restoration Framework.}
    \label{fig:mainfigure}
\vspace{-5pt}
\end{figure*}

\section{Our Method}

\subsection{Preliminaries}
\paragraph{Problem Definition} This paper investigates the missing modality scenario in multimodal learning. Without loss of generality, we study the input case involving $M=2$ modalities, denoted as $m_1$ and $m_2$ (text and image). Specifically, the multimodal dataset during the inference stage can be represented as:
$D = \{D_c, D_{m_1}, D_{m_2}\}$
where the subsets are defined as follows: The complete-modality subset $D_c = \{(x_{m_1}, x_{m_2}, y)\}$ includes inputs $x$ from all modalities and their corresponding labels $y$. The missing-modality subsets:
$D_{m_1} = \{(x_{m_1}, y)\}, \quad D_{m_2} = \{(x_{m_2}, y)\}$ represent the scenarios where one of the modalities is missing (missing text or missing image).

\paragraph{Diffusion Model} To capture complex high-dimensional distributions, the forward process gradually perturbs the initial VLM feature distribution $x_0 \sim q(x_0)$ by injecting Gaussian noise according to a variance scheduler $\{\beta_t\}_{t=1}^T$. 
Let $\alpha_t = 1 - \beta_t$ and $\bar{\alpha}_t = \prod_{i=1}^t \alpha_i$; the latent representation at any time step $t$ can be expressed as:
\begin{equation}
    q(x_t | x_0) = \mathcal{N}(x_t; \sqrt{\bar{\alpha}_t} x_0, (1 - \bar{\alpha}_t) \mathbf{I})
\end{equation}

The missing modality feature is reconstructed by iteratively reversing the diffusion process from $t=T$ down to $1$. 
Starting from Gaussian noise $x_T \sim \mathcal{N}(0, \mathbf{I})$, each intermediate latent state $x_{t-1}$ is derived from its successor $x_t$ through the following transition:
\begin{multline}
    x_{t-1} = \sqrt{\bar{\alpha}_{t-1}} \left( \frac{x_t - \sqrt{1-\bar{\alpha}_t} \epsilon_\theta(x_t, t \mid x_{m_1})}{\sqrt{\bar{\alpha}_t}} \right) \\
    + \sqrt{1-\bar{\alpha}_{t-1}} \epsilon_\theta(x_t, t \mid x_{m_1})
\end{multline}

where $\epsilon_\theta$ denotes the noise prediction network conditioned on the available modality feature $x_{m_1}$. 
This formulation ensures a stable restoration within the VLM's latent space.

\paragraph{Our Framework} To address missing modalities, the framework first employs a pre-trained foundational VLM to extract latent features from the available modality: $z_{m_1} = \text{Encoder}(x_{m_1})$. 
Conditioned on $z_{m_1}$, the missing feature $\hat{z}_{m_2}$ (denoted as $z_0$ in the diffusion process) is reconstructed via a $T$-step iterative DDIM sampling process. 
Finally, the available and restored features are concatenated and fed into a lightweight decoder to perform the downstream multimodal task: $y = \text{Decoder}([z_{m_1}; \hat{z}_{m_2}])$.


\subsection{Dynamic Modality Gating}

In the context of missing modality generation, the conventional structure of attention mechanisms and FFN in deep Transformers magnify several limitations:
\textbf{Low Feature-Background Discrimination:} Standard attention often fails to decouple critical semantic features from irrelevant background noise, which becomes particularly important when one modality is absent and the model must rely on limited information.
\textbf{Information Dilution:} In long-sequence processing, attention weights are distributed over numerous tokens, weakening the representation intensity of key information.
\textbf{Lack of Adaptive Control:} There is no inherent mechanism to dynamically modulate the focus and intensity of modality interactions.

To overcome these limitations, we integrate a modality gating unit into the Diffusion Transformer block (containing a self-attention layer, a cross-attention layer and a FFN) to dynamically guide the feature restoration process. We replace the direct connection with a dynamic gating mechanism conditioned on global semantics and the diffusion timestep. Specifically, we first fuse the timestep embedding $t_{emb}$ into the condition $C$: $ C = Condition_{emb} + \text{MLP}_{time}(t_{emb})$.
Subsequently, an attention pooling is utilized, where query probes (serving as learnable embedding weights) aggregate key semantics from the fused features:
\begin{equation}
    G_{pooled} = \text{Softmax}\left(\frac{Q_{probe}(W_K C)^T}{\sqrt{d_k}}\right)(W_V C)
\end{equation}
A non-linear mapping then produces the channel-wise modulation vector $G$:
\begin{equation}
    G = \text{MLP}_{gate}(G_{pooled})
\end{equation}
Within each Transformer block, specialized activation coefficients for the self-attention and MLP layers are computed via linear projection and Sigmoid activation:
\begin{equation}
    Z_{attn} = \sigma(W_{attn} G + b_{attn}), \quad Z_{mlp} = \sigma(W_{mlp} G + b_{mlp})
\end{equation}
The gating unit adaptively regulates the information flow across feature channels using these weights:
\begin{equation}
    x_{mid} = x + Z_{attn} \odot \text{Self-Attention}(\text{LN}(x))
\end{equation}
\begin{equation}
    x_{out} = x_{mid} + Z_{mlp} \odot \text{MLP}(\text{LN}(x_{mid}))
\end{equation}
where $\odot$ denotes the element-wise product. 

This learnable gating mechanism acts as a secondary filter, enhancing the signal-to-noise ratio of critical features. By suppressing irrelevant information flow, it effectively preserves the most salient cross-modal representations, providing robust guidance.

\subsection{Mutual Learning Mechanism}

We introduce a diffusion-based Mutual Learning framework that replaces isolated modality mappings with a bidirectional closed-loop system. By explicitly aligning the semantic spaces of the I2T and T2I models, our method prevents semantic drift and improves reconstruction quality under high cross-modal uncertainty.

During the standard training procedure, the model learns uni-directional mappings by minimizing the noise estimation error. For a given clean feature $x_0$ and conditional feature $c$, at a random timestep $t \sim [0, T]$, a noisy sample $x_t$ is generated through the forward diffusion process. The base training loss $\mathcal{L}_{base}$ is defined as the mean squared error between the predicted noise and the ground-truth Gaussian noise $\epsilon$: 

\vspace{-5pt}
\begin{equation}
\mathcal{L}_{base} = \mathbb{E}_{x_0, \epsilon, t} \left[ \| \epsilon - \epsilon_\theta(x_t, c, t) \|^2 \right]
\end{equation}
\vspace{-5pt}

To realize the ``predict-and-reconstruct'' closed loop, we introduce a novel training paradigm. When the diffusion timestep $t$ is below a preset threshold $\tau$ , the system is in a critical stage of denoising, where the quality of features generated by the model is vital for semantic alignment. First, the estimated value of the clean feature $\hat{x}_0$ is derived from the noisy sample $x_t$ using the noise $\epsilon_\theta$ predicted by the model:

\vspace{-5pt}
\begin{equation}
\hat{x}_0 = \frac{1}{\sqrt{\bar{\alpha}_t}} \,x_t - \sqrt{\frac{1}{\bar{\alpha}_t} - 1} \,\epsilon_\theta(x_t, c, t)
\end{equation}
\vspace{-5pt}

The mutual learning consistency loss $\mathcal{L}_{mutual}$ requires that the restored feature estimate $\hat{x}_0$ serves as an effective condition to guide the reverse-mapping model. For instance, the restored image feature $\hat{x}_{0,img}$ must be able to assist the text diffusion model in accurately predicting the text modality noise $\epsilon_{txt}$. The bi-directional mutual learning consistency loss function is defined as:
\vspace{-5pt}
\begin{multline}
\mathcal{L}_{mutual} = \mathbb{E} \Big[ \| \epsilon_{txt} - \epsilon_{\theta, i2t}(x_{t,txt}, \hat{x}_{0,img}, t) \|^2 \\
+ \| \epsilon_{img} - \epsilon_{\theta, t2i}(x_{t,img}, \hat{x}_{0,txt}, t) \|^2 \Big]
\end{multline}
where the former term characterizes the I2T consistency loss that evaluates the conditional accuracy of restored image features and the latter term represents the T2I consistency loss. In this way, the restored features are forced to be in the intelligible space of the other modality.

It is worth noting that we found that if $\mathcal{L}_{mutual}$ is used as the sole loss, the model tends to learn an identity mapping, neglecting the reconstruction accuracy of the original latent features. The final loss function $\mathcal{L}_{total}$ integrates the mutual learning loss and the base mapping regularization loss:
\begin{equation}
\mathcal{L}_{total} = \mathcal{L}_{mutual} + \mathcal{L}_{base, i2t} + \mathcal{L}_{base, t2i}
\end{equation}
Through joint training, the I2T and T2I models serve as mutual regularizers for each other. This explicit alignment mechanism effectively solves the problem of cross-modal semantic misalignment in uni-directional diffusion.

\subsection{Restorative Mid-Stage Training}

To further bolster the model’s resilience against missing modalities, we introduce a restorative mid-stage training stage after VLM’s standard contrastive alignment. We train two independent DiT models for the I2T and T2I restoration tasks, respectively. The DiT-based architecture allows the framework to handle various data scales and task requirements. Training is conducted on high-quality, large-scale benchmarks like CC3M and COCO. This strategic choice ensures the acquisition of robust, generalizable semantic embeddings that effectively span the comprehensive latent manifold established by VLM. Our overview framework is shown in Figure~\ref{fig:mainfigure}.

\definecolor{lightgray}{gray}{0.92}
\definecolor{mygreen}{RGB}{0, 110, 0}
\definecolor{myblue}{RGB}{244, 247, 254}
\definecolor{myyellow}{RGB}{255, 239, 243}

\begin{table*}[t] 
\centering

\caption{Performance comparisons of our method with various baseline methods on four datasets under 70\% missing rate: Our method achieves state-of-the-art (SOTA) across the board. This experiment uses a Zero-Shot mode of our 20-layer diffusion model pre-trained on the CC3M dataset. Bold denotes the best results and underline denotes the second-best. }
\vspace{-5pt}

\setlength{\extrarowheight}{0.1pt} 
\renewcommand{\arraystretch}{1.05} 

\resizebox{\textwidth}{!}{
\begin{tabular}{>{\raggedright\arraybackslash}p{2.58cm}>{\centering\arraybackslash}p{1.725cm}cccccccccccc}
\toprule \toprule

\textbf{Dataset} & & \multicolumn{3}{c}{\textbf{MM-IMDb}} & \multicolumn{3}{c}{\textbf{N24News}} & \multicolumn{3}{c}{\textbf{MMHS11K}} & \multicolumn{3}{c}{\textbf{Food101}} \\ 
\cmidrule(lr){3-5} \cmidrule(lr){6-8} \cmidrule(lr){9-11} \cmidrule(lr){12-14}

\textbf{Missing Type} & \textbf{Backbone} & Image & Text & Both & Image & Text & Both & Image & Text & Both & Image & Text & Both \\ 
\cmidrule(lr){3-3} \cmidrule(lr){4-4} \cmidrule(lr){5-5} \cmidrule(lr){6-6} \cmidrule(lr){7-7} \cmidrule(lr){8-8} \cmidrule(lr){9-9} \cmidrule(lr){10-10} \cmidrule(lr){11-11} \cmidrule(lr){12-12} \cmidrule(lr){13-13} \cmidrule(lr){14-14}

\textbf{Method} & & F1-M & F1-M & F1-M & ACC & ACC & ACC & ACC & ACC & ACC & ACC & ACC & ACC \\ \midrule[1pt] 


ShaSpec \color{gray}{(\textit{CVPR'23})} & - & 35.33 & 44.98 & 38.11 & 61.23 & 54.48 & 57.74 & 72.35 & 70.06 & 71.39 & 74.64 & 61.52 & 70.04 \\
MAP \color{gray}{(\textit{CVPR'23})} & ViLT & 38.74 & 47.44 & 39.13 & 61.34 & 57.67 & 57.78 & 78.28 & 72.59 & \underline{75.73} & 86.72 & 74.16 & 78.42 \\
RebQ \color{gray}{(\textit{ArXiv'24})} & ViLT & 18.52 & 22.40 & 19.85 & 52.47 & 48.91 & 50.12 & 58.60 & 56.44 & 57.25 & 71.24 & 68.57 & 70.01 \\
RobustPT \color{gray}{(\textit{ICMR'25})} & ViLT & 48.21 & 47.45 & 47.57 & 65.53 & 58.20 & 61.45 & 76.85 & 74.17 & 75.24 & 86.90 & 76.34 & 77.87 \\
RAGPT \color{gray}{(\textit{AAAI'25})} & ViLT & 38.82 & 54.33 & 49.55 & 61.18 & 57.14 & 60.81 & 76.93 & 73.18 & 74.94 & 82.42 & 76.51 & 77.54 \\
REDEEM \color{gray}{(\textit{KDD'25})} & ViLT & 47.20 & \underline{56.32} & 50.52 & 63.90 & 59.17 & 61.40 & \underline{78.31} & 71.66 & 74.67 & 83.78 & 77.52 & 78.77 \\
MDP \color{gray}{(\textit{IJCAI'25})} & ViLT & 48.61 & 40.36 & 41.62 & 62.13 & 58.29 & 59.20 & 74.36 & \underline{74.22} & 74.12 & 87.07 & 75.00 & 79.62 \\
KB \color{gray}{(\textit{CVPR'25})} & Qwen-VL-2B & 48.37 & 51.25 & 50.20 & 63.45 & 56.71 & 60.33 & 75.63 & 72.69 & 74.07 & 85.67 & 74.59 & 79.81 \\
KB \color{gray}{(\textit{CVPR'25})} & Qwen-VL-7B & 51.21 & 52.51 & 51.45 & 63.87 & 57.11 & 61.02 & 76.67 & 73.25 & 74.97 & 86.42 & 75.68 & 79.99 \\
DCP \color{gray}{(\textit{NeurIPS'24})} & CLIP ViT-B16 & 51.15 & 47.58 & 47.47 & 67.82 & 57.83 & 61.82 & 71.66 & 73.64 & 70.38 & 87.74 & 77.82 & 81.83 \\
SyP \color{gray}{(\textit{ICCV'25})} & CLIP ViT-B16 & 50.16 & 46.52 & 48.02 & 67.73 & 57.64 & 61.73 & 70.73 & 73.87 & 70.85 & 87.82 & 78.01 & 81.27 \\ 
MoRA \color{gray}{(\textit{ICLR'26})} & CLIP ViT-B16 & \underline{52.40} & 50.72 & \underline{51.61} & \underline{68.77} & \underline{59.38} & \underline{63.35} & 77.79 & 73.87 & 74.94 & \underline{88.02} & \underline{78.32} & \underline{82.26} \\

\rowcolor{myblue} \textbf{Ours} & CLIP ViT-B32 & \textbf{58.22} & \textbf{57.24} & \textbf{57.49} & \textbf{71.70} & \textbf{61.89} & \textbf{65.97} & \textbf{86.47} & \textbf{79.55} & \textbf{82.00} & \textbf{88.37} & \textbf{78.71} & \textbf{83.01} \\ 


\rowcolor{myyellow} \textbf{Ours} & CLIP ViT-B16 & \textbf{58.74} & \textbf{57.85} & \textbf{57.91} & \textbf{72.81} & \textbf{63.17} & \textbf{66.58} & \textbf{87.65} & \textbf{80.82} & \textbf{83.04} & \textbf{89.14} & \textbf{81.89} & \textbf{84.11} \\

Improve (\%) & & \color{mygreen}{\textbf{+7.53}} & \color{mygreen}{\textbf{+1.53}} & \color{mygreen}{\textbf{+6.30}} & \color{mygreen}{\textbf{+4.04}} & \color{mygreen}{\textbf{+3.79}} & \color{mygreen}{\textbf{+3.23}} & \color{mygreen}{\textbf{+9.34}} & \color{mygreen}{\textbf{+6.60}} & \color{mygreen}{\textbf{+7.31}} & \color{mygreen}{\textbf{+1.12}} & \color{mygreen}{\textbf{+3.57}} & \color{mygreen}{\textbf{+1.85}} \\ \bottomrule \bottomrule
\end{tabular}
}

\label{tab:main}
\end{table*}

\section{Experiments}

\subsection{Datasets and Evaluation Metrics}

\subsubsection{Datasets} Following prior research, we evaluated our method on four representative multimodal datasets: (1)\emph{MM-IMDb}~\cite{arevalo2017gated}: Primarily used for multi-label movie classification. (2)\emph{N24News}~\cite{wang-EtAl:2022:LREC3}: Designed for multi-modal news classification. (3)\emph{MMHS11K}~\cite{saddozai2025multimodal}: Focused on multi-modal hate speech detection. (4)\emph{Food101}~\cite{bossard14}: Focused on food classification. All dataset splits were kept consistent with the original publications, involving image and text modalities.

\subsubsection{Evaluation Metrics} Following established protocols in prior literature, we employ two metrics for each dataset: F1-Macro (F1-M) is utilized for the \emph{MM-IMDb} dataset, while Accuracy (ACC) is adopted for the \emph{Food101}, \emph{N24News} and \emph{MMHS11K} datasets.

\subsubsection{Missing Pattern Settings} 
The missing rate $\eta\%$ is defined as the fraction of incomplete samples within the total dataset. Specifically, we investigate the following configurations:
Single Modality Missing: Given a missing rate $\eta\%$, a subset of $\eta\%$ of the total samples is restricted to a single modality, while the remaining $(1-\eta)\%$ consists of complete multimodal pairs.
Both Modalities Missing: The missing modalities are distributed equally across both domains. $\frac{\eta}{2}\%$ of the dataset consists of image-only samples, $\frac{\eta}{2}\%$ contains only text, and the remaining $(1-\eta)\%$ of the data remains complete with both modalities. For fairness, all experiments share identical data splits across both missing and complete modality settings.

\subsection{Implementation Details}

Following prior research, we employ a pre-trained CLIP~\cite{radford2021learning} as the frozen VLM network and utilize large-scale multimodal dataset CC3M~\cite{sharma2018conceptual} as our mid-stage training datasets. A detailed analysis regarding the restorative mid-stage training dataset can be found in the subsequent sections. For downstream tasks, we train an extremely lightweight decoder consisting of a transformer with only two self-attention layers and one cross-attention layer, along with a classifier. Since the DiT’s pre-training and downstream datasets are disjoint, the module is kept frozen to operate in a purely zero-shot manner, effectively decoupling the general feature restoration capability from task-specific label learning.


For restorative mid-stage training, the DiT ($\beta=1 \times 10^{-4}$) is optimized for 40 epochs using an AdamW optimizer with a $lr = 1 \times 10^{-4}$ and Mixed-Precision (BF16) on an NVIDIA A800 GPU. For downstream tasks, the decoder is trained for 20 epochs on a single NVIDIA RTX A5000 GPU (24GB) using the AdamW optimizer, with a $lr = 5 \times 10^{-4}$ that decays by a factor of 0.1 at the midpoint. 

For inference, the diffusion model employs DDIM sampling with 50 steps; for a detailed analysis of the sampling steps, please refer to the subsequent sections. In the Mutual Learning Mechanism, we set the coefficient of the threshold to $\tau = 50$. 
\subsection{Main Results}

To verify the effectiveness of our method and its Zero-Shot feature restoration capability, we evaluate the frozen diffusion module against $12$ highly competitive baseline models across $4$ datasets with a missing rate $\eta\% = 70\%$. From results in Table~\ref{tab:main}: Under various missing modality conditions, Under various missing modality conditions, our model comprehensively outperforms baseline methods, including backbone-equivalent counterparts, and even exceeds Qwen-VL, despite the latter possessing 15 times the parameter scale. 
Across all datasets, our method achieves an average performance gain of $\mathbf{4.68\%}$, with a maximum improvement of $\mathbf{9.34\%}$.

Notably, during downstream task testing, our method completely freezes the enhanced diffusion module and only trains decoders. Our pre-trained diffusion module achieves SOTA performance through zero-shot feature restoration on unseen datasets. 
This demonstrates its robust generalization in modality-missing scenarios while preserving the core capabilities of the foundation VLM.

Conversely, imputation and retrieval-based methods, such as SMIL~\cite{ma2021smil}, ShaSpec~\cite{wang2023multi}, RebQ~\cite{zhao2024reconstruct}, RAGPT~\cite{lang2025retrieval}, REDEEM~\cite{lang2025redeeming}, Knowledge Bridger~\cite{ke2025knowledge}, show suboptimal performance. This stems from heuristic padding noise and the inherent semantic gap during retrieval, which, compounded by modality heterogeneity, create significant reconstruction bottlenecks, especially when retrieved references deviate from the target distribution. Besides, MoRA ~\cite{zhao2025mora} employs a parameter-efficient fine-tuning method to VLM.

Furthermore, prompt-based methods exhibit limited effectiveness in missing-modality scenarios. Their adaptability is restricted either by static prompt strategies, such as MAP~\cite{lee2023multimodal}, MDP~\cite{zhao2025enhancing}, and DCP~\cite{shi2024deep}, or by dynamic prompting approaches like RobustPT~\cite{dai2025robustpt} and Synergistic Prompting~\cite{zhang2025synergistic} that, despite attempting cross-modal alignment, fail to reconstruct explicit features.  Additionally, prompt-based methods may impose constraints on application scenarios, as they cannot be applied to tasks that require complete, high-fidelity features for decoding, a gap that our diffusion-based restoration approach effectively bridges.

\subsection{Ablation Study}

\begin{figure}[t]
    \centering
    \includegraphics[width=0.48\textwidth, trim=0cm 0cm 0cm 0cm, clip]{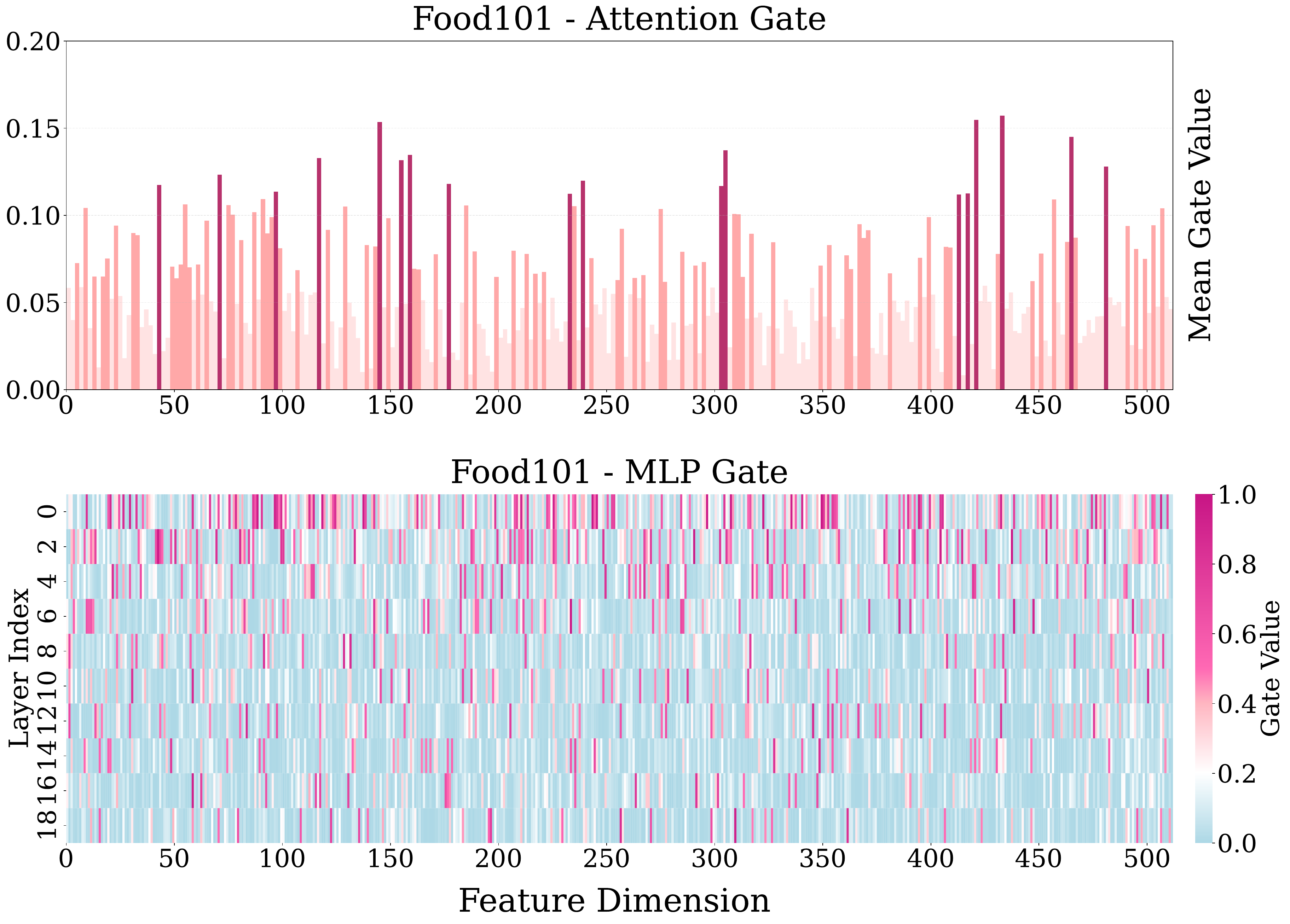}
    \vspace{-15pt}
    \caption{The visualization of Dynamic Modality Gating: The top plot shows the average activation values of each channel after attention-based gating. The bottom plot shows the activation values of each channel at different DiT depths.}
    \Description{The visualization of Dynamic Modality Gating.}
    \label{fig:gate_analysis}
\vspace{-3pt}
\end{figure}

\subsubsection{Ablation of Dynamic Modality Gating} To analyze the impact of Dynamic Modality Gating in our improved Diffusion Transformer, we select two variants originally proposed in Diffusion models and widely applied in image generation tasks, along with a baseline: (1) Adaptive LayerNorm (AdaLN): Utilizing AdaLN instead of our gating mechanism to achieve linear scaling of features; (2) Concat: Simply concatenating conditional features with original features and appending an MLP to scale the feature dimensions; (3) Baseline: Completely removing the gating mechanism while retaining only the basic DiT framework~\cite{peebles2023scalable}. As shown in Table~\ref{tab:ablation}, our ablation study demonstrate that the Modality Gating mechanism achieves the superior performance. We attribute its effectiveness to its ability to filter components containing critical semantic information while simultaneously suppressing modality-irrelevant noise in missing-modality tasks, thereby achieving precise control over the output of the diffusion model.

\subsubsection{Ablation of Mutual Learning Mechanism} To evaluate the impact of the Mutual Learning Mechanism on the Diffusion model, we designed the following experiments: (1) With Mutual Learning Mechanism: Applying Mutual Learning; (2) Without Mutual Learning : Completely removing the Mutual Learning Mechanism and conducting only a single-stage training. We observe that the Mutual Learning Mechanism improves performance (Table~\ref{tab:ablation}). This finding supports our assertion that the Mutual Learning Mechanism effectively addresses the issue of semantic drift in unidirectional diffusion, ensuring that the features generated by the model possess bidirectional interchangeability beneficial for downstream tasks.

\subsubsection{Effect of Dynamic Modality Gating} To further investigate the essence of our Gating and explore whether the model can automatically prune meaningless components of the output, we perform a visualization analysis of the gates, as illustrated in Figure~\ref{fig:gate_analysis}. The results reveal distinct sparsity in the gates following the attention, proving that the gating mechanism effectively functions as a filter. Observations of the gates following the MLP indicate that the overall activation intensity gradually decreases as the depth of the network  increases. This suggests that in shallower DiT modules, the model needs to incorporate more information from conditional features, thereby requiring more channels to preserve initialization information. Conversely, in deeper DiT modules, the model prioritizes the selection of critical semantic information while subtracting more irrelevant noise, resulting in a lower overall activation intensity across most channels. During the training process, we observed that the mean value of the gates at the output is approximately $0.1$, with the vast majority of values falling within the range $0-0.2$ . The model automatically optimizes the gates to become zero-valued for most entries while magnifying only the essential features. Overall, our modality gating evolves into a sparse ``selector''.

\definecolor{lightgray}{gray}{0.9}
\definecolor{DarkRed}{rgb}{0.6, 0, 0}
\definecolor{DarkBlue}{RGB}{3, 38, 139}
\begin{table}[t]
\centering

\caption{Ablation Study of different modules on \emph{MMHS11K} and \emph{N24News} with CLIP ViT-B32 Backbone and 20 DiT blocks.}
\vspace{-5pt}
\renewcommand{\arraystretch}{1.1} 
\setlength{\tabcolsep}{3pt} 

\resizebox{\columnwidth}{!}{
\begin{tabular}{lcccccc}
\specialrule{1.1pt}{0pt}{2.5pt}
\multirow{2}{*}{\textbf{Method}} & \multicolumn{3}{c}{\textbf{\emph{MMHS11K}}} & \multicolumn{3}{c}{\textbf{\emph{N24News}}} \\ \cmidrule(lr){2-4} \cmidrule(lr){5-7}
& Image & Text & Both & Image & Text & Both \\ \midrule

\textbf{\textcolor{black}{Ours (All)}} & \textbf{\textcolor{DarkRed}{86.47}} & \textbf{\textcolor{DarkRed}{79.55}} & \textbf{\textcolor{DarkRed}{82.00}} & \textbf{\textcolor{DarkRed}{71.70}} & \textbf{\textcolor{DarkRed}{61.89}} & \textbf{\textcolor{DarkRed}{65.97}} \\ \midrule

\textbf{Mutual} & \textcolor{DarkBlue}{\underline{85.27}} & \textcolor{DarkBlue}{\underline{76.94}} & \textcolor{DarkBlue}{\underline{80.79}} & \textcolor{DarkBlue}{\underline{71.37}} & \textcolor{DarkBlue}{\underline{61.38}} & \textcolor{DarkBlue}{\underline{65.15}} \\
Base  & 84.06 & 76.23 & 80.11 & 70.18 & 60.05 & 63.37 \\ \midrule

\textbf{Gating} & \textcolor{DarkBlue}{\underline{85.64}} & \textcolor{DarkBlue}{\underline{78.00}} & \textcolor{DarkBlue}{\underline{80.77}} & \textcolor{DarkBlue}{\underline{71.38}} & \textcolor{DarkBlue}{\underline{61.48}} & \textcolor{DarkBlue}{\underline{64.93}} \\
AdaLN & 84.44 & 77.33 & 80.64 & 70.77 & 60.50 & 63.88 \\
Concat & 80.05 & 72.14 & 76.44 & 65.21 & 54.95 & 59.63 \\ 
Base & 84.06 & 76.23 & 80.11 & 70.18 & 60.05 & 63.37 \\

\specialrule{1.1pt}{1.5pt}{0pt}
\end{tabular}
}

\label{tab:ablation}
\vspace{2pt}
\end{table}

\definecolor{lightgray}{gray}{0.9}

\begin{table}[t] 
\centering
\caption{Evaluation across different sizes of mid-stage training datasets with CLIP ViT-B32 Backbone and 20 DiT blocks.}
\vspace{-5pt}
\label{tab:pretraining_impact}
\renewcommand{\arraystretch}{1.1} 
\setlength{\tabcolsep}{2.5pt} 

\resizebox{\columnwidth}{!}{ 
\begin{tabular}{lccccccc}
\specialrule{1.0pt}{0pt}{2.5pt}
\multirow{2}{*}{\textbf{Pretraining}} & \multirow{2}{*}{\textbf{Size}} & \multicolumn{3}{c}{\textbf{\emph{MM-IMDb}}} & \multicolumn{3}{c}{\textbf{\emph{MMHS11K}}} \\ \cmidrule(lr){3-5} \cmidrule(lr){6-8}
& & Image & Text & Both & Image & Text & Both \\ \midrule

\rowcolor{lightgray}
\textbf{CC3M (Ours)} & 3M & \textbf{58.22} & \textbf{57.24} & \textbf{57.49} & \textbf{86.47} & \textbf{79.55} & \textbf{82.00} \\
COCO+Flickr30k & 475K & 57.57 & 57.12 & 57.19 & 85.59 & 78.77 & 81.23 \\
COCO & 445K & 57.28 & 56.95 & 56.84 & 85.23 & 78.69 & 80.95 \\
Visual Genome & 108K & 57.00 & 56.41 & 56.47 & 83.98 & 76.99 & 78.91 \\
Food101 & 68K & 56.79 & 55.84 & 55.97 & 83.79 & 76.21 & 78.34 \\ \midrule

Baseline (DCP) & - & 51.15 & 47.58 & 47.47 & 71.66 & 73.64 & 70.38 \\ 
\specialrule{1.0pt}{1.5pt}{0pt}
\end{tabular}
}
\vspace{2pt}
\end{table}

\definecolor{lightgray}{gray}{0.9}

\begin{table}[t]
\centering

\caption{Performance comparison with various DiT depths (with CLIP ViT-B32 Backbone and CC3M dataset).}
\vspace{-5pt}
\label{tab:DiT_depth_analysis}

\renewcommand{\arraystretch}{1.1} 
\setlength{\tabcolsep}{2pt} 

\resizebox{\columnwidth}{!}{
\begin{tabular}{lccccccccc}
\specialrule{1.0pt}{0pt}{2.5pt}
\multirow{2}{*}{\textbf{DiT Depth}} & \multicolumn{3}{c}{\textbf{\emph{MM-IMDb}}} & \multicolumn{3}{c}{\textbf{\emph{N24News}}} & \multicolumn{3}{c}{\textbf{\emph{MMHS11K}}} \\ \cmidrule(lr){2-4} \cmidrule(lr){5-7} \cmidrule(lr){8-10}
& Image & Text & Both & Image & Text & Both & Image & Text & Both \\ \midrule

Depth = 16 & 57.37 & 56.20 & 57.01 & 71.16 & 61.39 & 65.31 & 85.79 & 79.61 & 81.20 \\

Depth = 18 & 57.79 & 56.46 & 57.20 & 71.24 & 61.46 & 65.33 & 85.82 & \textbf{79.59} & 81.27 \\

\rowcolor{lightgray} 
\textbf{Depth = 20} & \textbf{58.22} & 57.24 & \textbf{57.49} & 71.70 & 61.89 & \textbf{65.97} & \textbf{86.47} & 79.55 & 82.00 \\

Depth = 22 & 57.38 & \textbf{57.78} & 56.93 & \textbf{72.01} & 61.76 & 65.77 & 85.55 & 79.45 & 81.68 \\

Depth = 24 & 57.18 & 56.99 & 56.42 & 71.11 & \textbf{61.97} & 65.71 & 85.05 & 78.36 & \textbf{82.41} \\ 
\specialrule{1.0pt}{1.5pt}{0pt}
\end{tabular}
}
\end{table}

\subsection{Model Scalability}

\definecolor{mygreen}{RGB}{0, 120, 0}

\begin{table*}[t]
\centering
\caption{Performance comparison validating the effectiveness of integrating our DiT module with DCP and SyP baselines. Experiments are conducted on three downstream datasets (\emph{MM-IMDb}, \emph{N24News} and \emph{MMHS11K}) with CLIP ViT-B16 Backbone.}
\label{tab:performance_comparison}

\renewcommand{\arraystretch}{0.9} 

\resizebox{\textwidth}{!}{%
\begin{tabular}{lccccccccc}
\toprule
\multirow{2}{*}{\textbf{Method}} & \multicolumn{3}{c}{\textbf{\emph{MM-IMDb}}} & \multicolumn{3}{c}{\textbf{\emph{N24News}}} & \multicolumn{3}{c}{\textbf{\emph{MMHS11K}} } \\ 
\cmidrule(lr){2-4} \cmidrule(lr){5-7} \cmidrule(lr){8-10} 
 & \textbf{Image} (70\%) & \textbf{Text} (70\%) & \textbf{Both} (70\%) & \textbf{Image} (70\%) & \textbf{Text} (70\%) & \textbf{Both} (70\%) & \textbf{Image} (70\%) & \textbf{Text} (70\%) & \textbf{Both} (70\%) \\ 
\midrule
DCP & 51.15 & 47.58 & 47.47 & 67.82 & 57.83 & 61.82 & 71.66 & 73.64 & 70.38 \\
DCP + DiT & \textbf{54.22}\textbf{\scriptsize\textcolor{mygreen}{(+3.07)}} & \textbf{49.96}\textbf{\scriptsize\textcolor{mygreen}{(+2.38)}} & \textbf{48.10}\textbf{\scriptsize\textcolor{mygreen}{(+0.63)}} & \textbf{71.56}\textbf{\scriptsize\textcolor{mygreen}{(+3.74)}} & \textbf{58.17}\textbf{\scriptsize\textcolor{mygreen}{(+0.34)}} & \textbf{62.64}\textbf{\scriptsize\textcolor{mygreen}{(+0.82)}} & \textbf{75.61}\textbf{\scriptsize\textcolor{mygreen}{(+3.95)}} & \textbf{74.22}\textbf{\scriptsize\textcolor{mygreen}{(+0.58)}} & \textbf{71.31}\textbf{\scriptsize\textcolor{mygreen}{(+0.93)}} \\
\midrule
SyP & 50.16 & 46.52 & 48.02 & 67.73 & 57.64 & 61.73 & 70.73 & 73.87 & 70.85 \\
SyP + DiT & \textbf{52.92}\textbf{\scriptsize\textcolor{mygreen}{(+2.76)}} & \textbf{46.89}\textbf{\scriptsize\textcolor{mygreen}{(+0.37)}} & \textbf{48.65}\textbf{\scriptsize\textcolor{mygreen}{(+0.63)}} & \textbf{71.46}\textbf{\scriptsize\textcolor{mygreen}{(+3.73)}} & \textbf{59.07}\textbf{\scriptsize\textcolor{mygreen}{(+1.43)}} & \textbf{62.55}\textbf{\scriptsize\textcolor{mygreen}{(+0.82)}} & \textbf{72.61}\textbf{\scriptsize\textcolor{mygreen}{(+1.88)}} & \textbf{75.45}\textbf{\scriptsize\textcolor{mygreen}{(+1.58)}} & \textbf{71.79}\textbf{\scriptsize\textcolor{mygreen}{(+0.94)}} \\ 
\bottomrule
\end{tabular}%
}
\end{table*}

\begin{figure*}[t]
    \centering
    \includegraphics[width=0.99\textwidth, trim=0cm 0cm 0cm 0.1cm, clip]{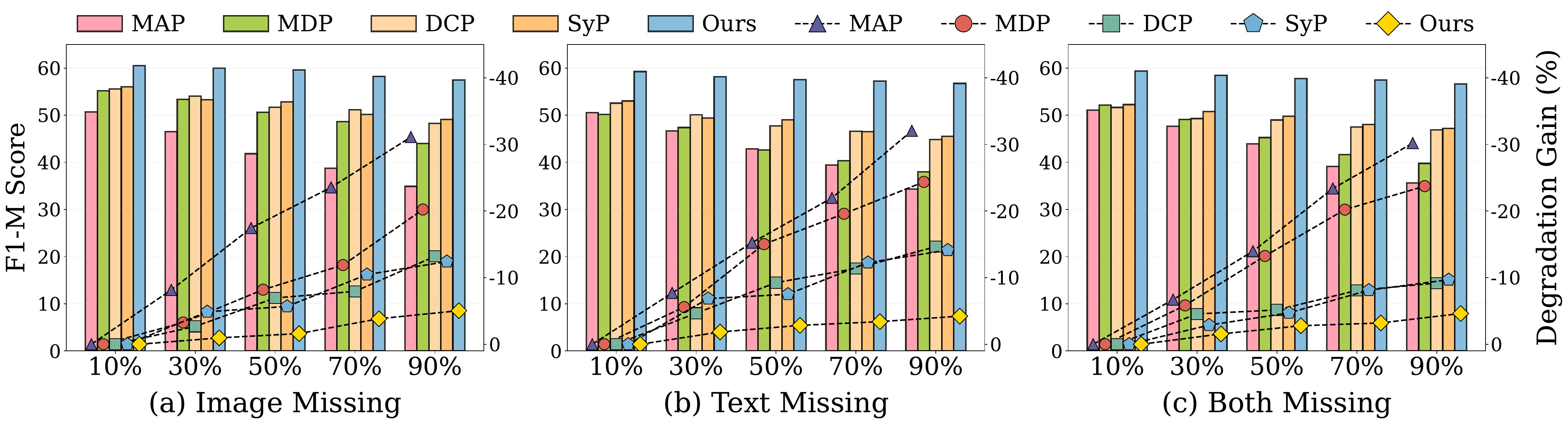}
    \vspace{-7pt}
    \caption{Robustness analysis on the \emph{MM-IMDb} dataset
across various missing rates in terms of F1-M Score.}
    \Description{Robustness analysis.}
    \label{fig:robustness_analysiss}
\end{figure*}

\subsubsection{Restorative Mid-Stage Training Datasets} To verify the scalability of our framework, we conducted mid-stage training on image-text datasets of varying scales and evaluated performance on downstream tasks (Table~\ref{tab:pretraining_impact}). First, the model achieves stable performance on \emph{MM-IMDb} and \emph{N24News} even when trained on small-scale datasets (e.g., \emph{Food101}~\cite{bossard14}). This indicates that the DiT-based architecture can capture semantic priors of the CLIP manifold with limited data. Second, downstream metrics exhibit a consistent upward trend as the training data scale expands from 0.44M (COCO~\cite{lin2014microsoft}) to 3M (CC3M~\cite{sharma2018conceptual}). For instance, the accuracy on \emph{MMHS11K} under the image-missing scenario improves from $85.59\%$ to $86.47\%$. These results validate that the framework scales effectively with increased data volume, leading to enhanced performance.

\subsubsection{Number of DiT Layers} To investigate the impact of model scale, we conducted an ablation study on the DiT architecture's depth (Table~\ref{tab:DiT_depth_analysis}). As depth increases from 16 to 20 layers, performance across downstream tasks consistently improves, indicating that deeper Transformer layers better model the complex CLIP manifold. However, further increasing depth to 22 or 24 layers yields diminishing returns, potentially due to overfitting as the model complexity increases while the dataset size remains constant. Consequently, we adopted a 20-layer DiT to optimally balance classification accuracy with computational overhead.

\subsection{Robustness to Different Missing Rates }

We conducted an experiment to analyze the robustness of our model against different missing rates. The Figure~\ref{fig:robustness_analysiss} illustrates the comparative results between our method and four strong baseline models (MAP, MDP, DCP, and SyP) on the \emph{MM-IMDb} dataset. We observe that the performance of all baseline models degrades significantly as the missing rate increases, with the degree of decline exceeding $10\%$ (across the range of $10\%$ to $90\%$ missing rates). In contrast, the performance of our method only decreases slightly with the increase in missing rate, showing a decline of only approximately $3\%$. This result highlights that our restorative mid-stage training module, serving as an extension of the base VLM, effectively restores features with consistent semantic information, thereby enhancing effectiveness under extremely high missing rates.


\subsection{Integration with Other Frameworks}

As a plug-and-play restorative mid-stage training module, our method offers high portability. Furthermore, since it does not alter the underlying VLM architecture, it is compatible and can be integrated with most existing missing modality approaches. The following parts analyze the broad applicability of our method and its performance when integrated with other frameworks.

To validate this scalability, we selected two representative prompt learning frameworks, Deep Correlated Prompting (DCP) and Synergistic Prompting (SyP), as experimental baselines. Standard implementations of these methods typically resort to static learnable prompts or zero-filling strategies to compensate for missing modalities. While effective to a degree, these non-generative approaches inherently struggle to recover instance-specific semantic details. We seamlessly integrated our proposed DiT module into their inference pipelines, replacing these static placeholders with generatively restored features.

As presented in Table~\ref{tab:performance_comparison}, quantitative results demonstrate that augmenting existing frameworks with our method yields consistent and significant performance gains across all missing scenarios. Specifically, on the \emph{MM-IMDb} dataset, the integrated \textbf{DCP+DiT} outperforms the original DCP by approximately \textbf{$3.07\%$} in F1-Macro under the Image-only condition (rising from $51.15\%$ to $54.22\%$), and by \textbf{$2.38\%$} under the Text-only condition. Similar trends are observed with \textbf{SyP+DiT} on the \emph{N24News} dataset, where the module boosts accuracy by \textbf{$3.73\%$} (from $67.73\%$ to $71.46\%$) in the Image-only scenario. These results strongly support our claim that the proposed module serves as a universal, performance-enhancing plugin that effectively bridges the semantic gap in existing systems without compromising their original architectural integrity.

\newcolumntype{Y}{>{\centering\arraybackslash}X}

\begin{table}[ht]
\centering
\caption{Performance Stability and Inference Efficiency across Sample Steps with CLIP ViT-B32 backbone.}
\label{tab:sample_steps_performance}
\vspace{-5pt}
\footnotesize 
\setlength{\tabcolsep}{2pt} 

\begin{tabularx}{\columnwidth}{@{} 
    >{\centering\arraybackslash}p{25pt}   
    >{\centering\arraybackslash}p{30pt}   
    >{\centering\arraybackslash}p{35pt}   
    *{6}{Y}                               
@{}} 
\toprule
\multirow{2}{*}{\textbf{Steps}} & \multicolumn{2}{c}{\textbf{Time (\emph{ms})}} & \multicolumn{3}{c}{\textbf{\emph{N24News}}} & \multicolumn{3}{c}{\textbf{\emph{MMHS11K}}} \\ \cmidrule(lr){2-3} \cmidrule(lr){4-6} \cmidrule(lr){7-9} 
 & \textbf{MissText} & \textbf{MissImage} & \textbf{Img} & \textbf{Txt} & \textbf{Both} & \textbf{Img} & \textbf{Txt} & \textbf{Both} \\ \midrule
\textbf{20}  & 296.8 & 274.5 & 71.26 & 61.32 & 65.56 & 85.95 & 79.07 & 81.45 \\
\textbf{30}  & 410.5 & 381.6 & 71.62 & 61.74 & 65.86 & 86.24 & 79.21 & 81.76 \\
\textbf{50}  & 707.4 & 678.7 & 71.70 & 61.89 & 65.97 & 86.47 & 79.55 & 82.00 \\
\textbf{100} & 1389.0 & 1338.1 & 71.81 & 61.89 & 65.99 & 86.50 & 79.56 & 82.17 \\
\textbf{200} & 2757.1 & 2737.0 & 71.86 & 61.97 & 66.01 & 86.54 & 79.56 & 82.23 \\
\textbf{500} & 6900.4 & 6747.7 & 72.02 & 62.00 & 66.07 & 86.55 & 79.58 & 82.31 \\ \bottomrule
\end{tabularx}
\end{table}

\subsection{Visualization of Model Effectiveness}

\subsubsection{T-SNE Analysis of Reconstructed Features} 
\begin{figure}[b]
\vspace{-10pt}
    \centering
    \includegraphics[width=0.48\textwidth, trim=0.5cm 0.5cm 0.5cm 0.5cm, clip]{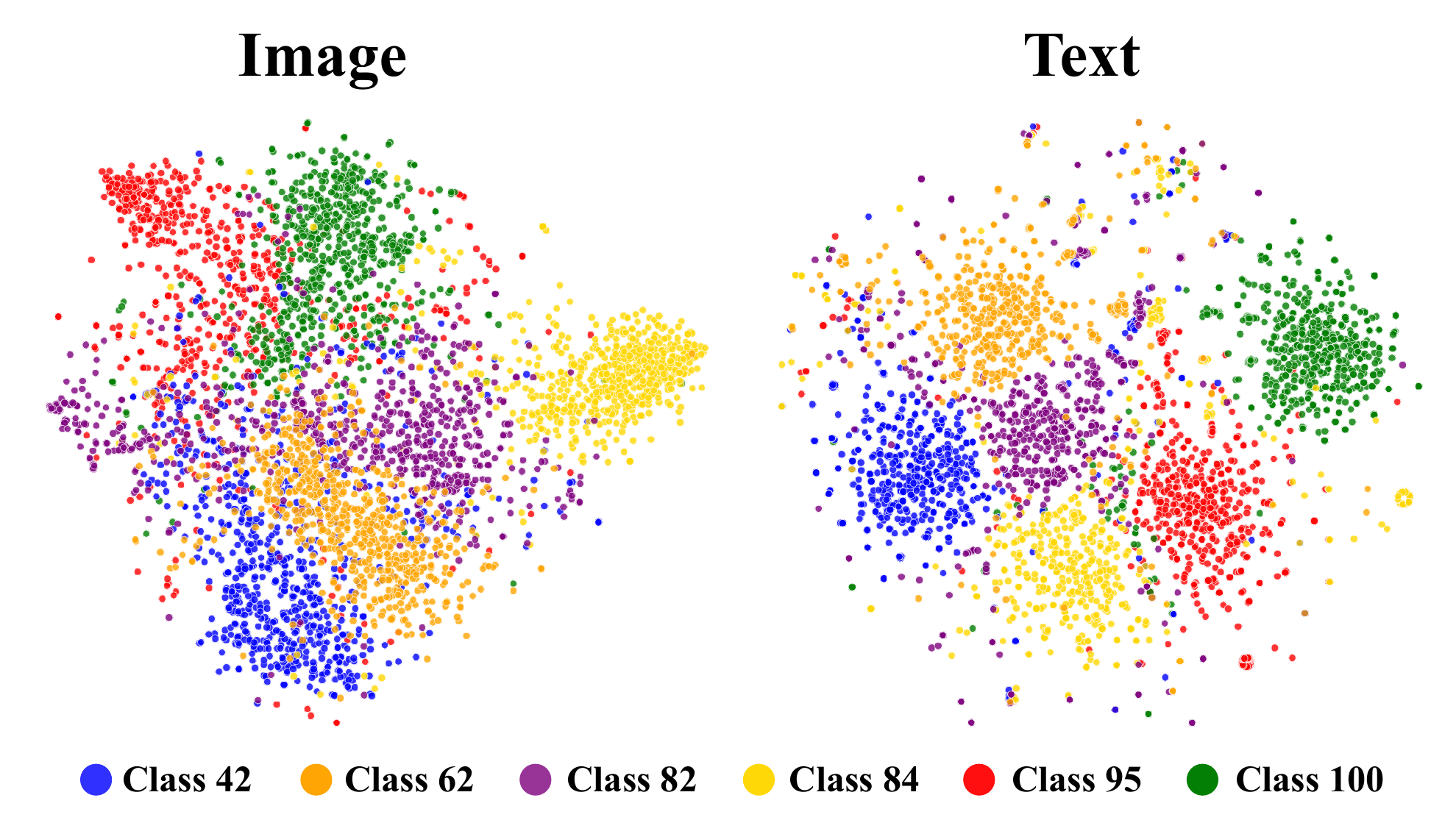}
    \caption{T-SNE analysis of restored features on the \emph{Food101} dataset, where we visualize the two-dimensional multimodal embedding distribution of specific categories.}
    \Description{T-SNE analysis of restored features.}
    \label{fig:tsne}
\end{figure}

To visually evaluate the semantic fidelity of the restored representations, we conduct a 2D t-SNE visualization of the restored image and text features, as shown in Figure~\ref{fig:tsne}. Qualitative analysis reveals that the restored features exhibit robust cluster cohesion and manifold separation within the 2D space.

\begin{figure}[t]
    \centering
    \includegraphics[width=0.48\textwidth, trim=0cm 0cm 0cm 0cm, clip]{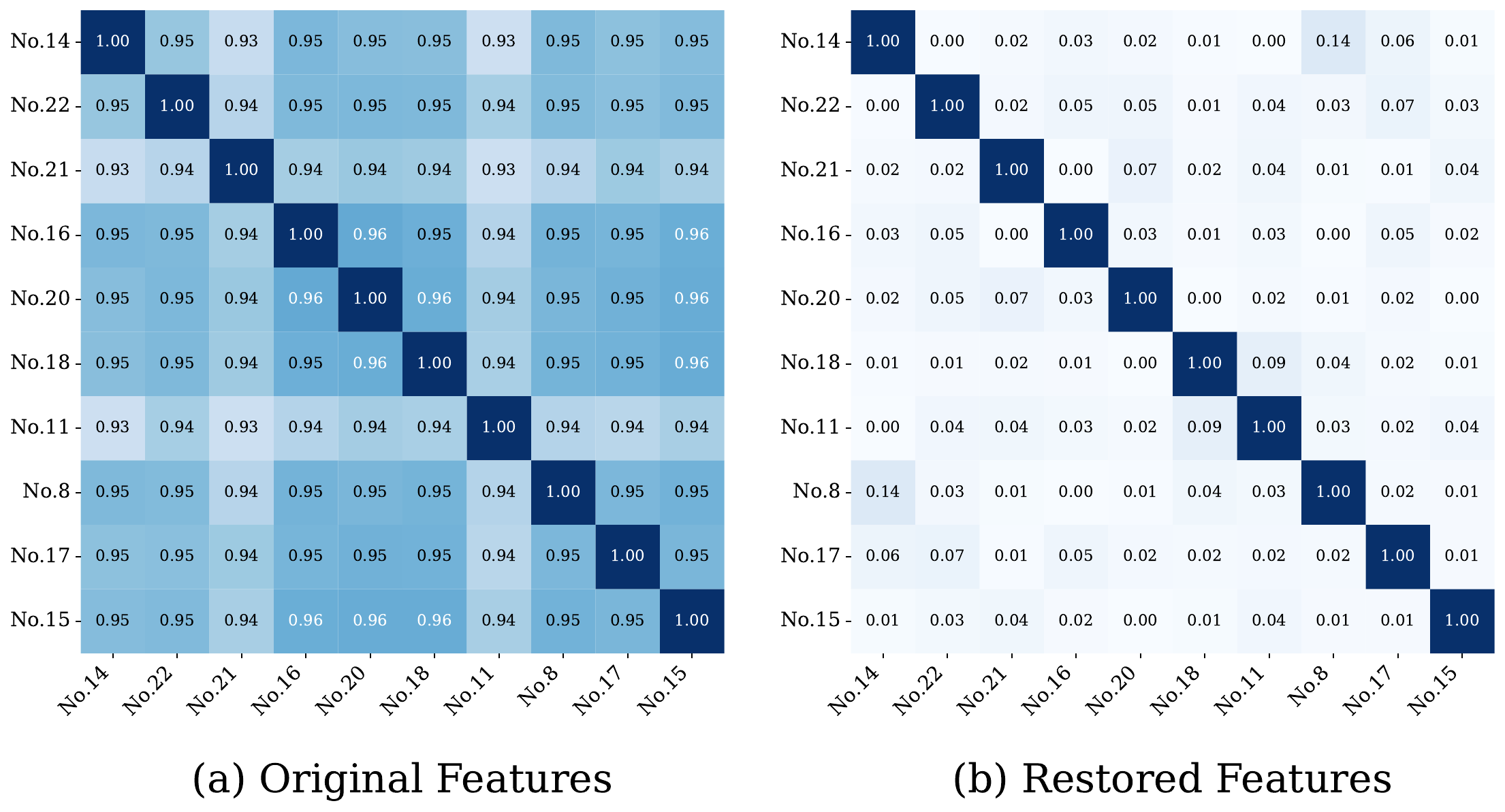}
    \vspace{-15pt}
    \caption{Category-wise cosine similarity for 1,000 features in \emph{MM-IMDb} datasets. The original CLIP features (left) and our restored counterparts (right) exhibit semantic alignment.}
    \Description{Category-wise cosine similarity.}
    \label{fig:category_similarity}
\end{figure}

\begin{figure}[t]
    \centering
    \includegraphics[width=0.48\textwidth, trim=0cm 0cm 0cm 0cm, clip]{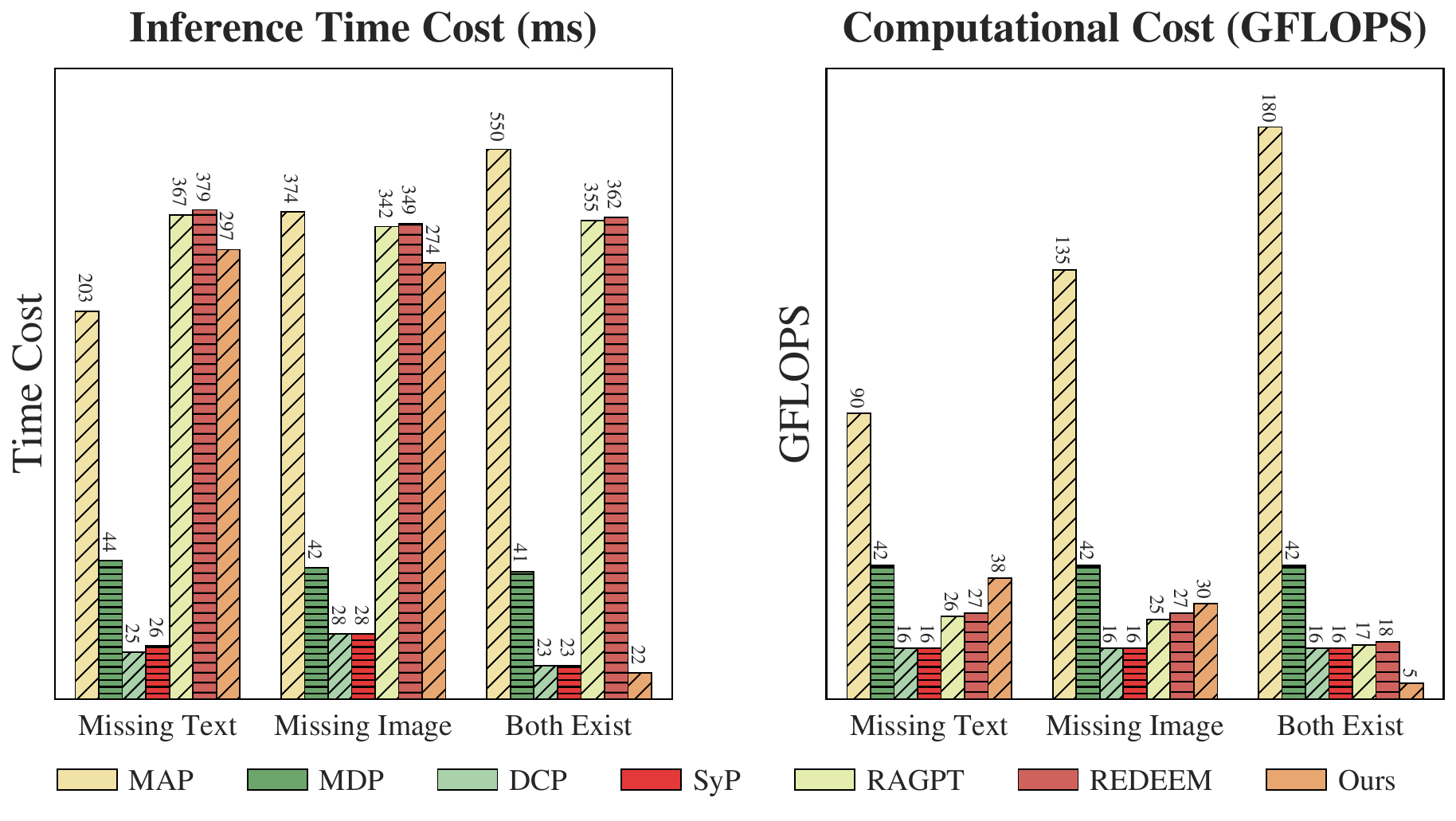}
    \vspace{-15pt}
    \caption{Comparison of inference efficiency and computational cost (GFLOPS) across baseline methods.}
    \Description{Comparison of inference efficiency and computational cost.}
    \label{fig:inference_efficiency}
\end{figure}

\subsubsection{Semantic Fidelity Analysis of Restored Features} To assess semantic fidelity, we compare original CLIP visual features $\mathbf{f}_{real}$ with our reconstructed features $\mathbf{f}_{recon}$ using inter-class similarity matrices across 10 \emph{MM-IMDb} categories. As shown in Figure~\ref{fig:category_similarity}, our restored features exhibit a more pronounced diagonal structure, signifying a transition from ``pixel-level reconstruction'' to ``manifold extraction.'' While CLIP features encapsulate pre-training redundancies and blurred semantic boundaries, our DiT-based diffusion process leverages text-guided class priors to enforce tighter constraints. This cross-modal mutual learning effectively mitigates modal noise, re-projecting features into highly discriminative clusters and reinforcing underlying category manifolds.

\subsection{Sampling and Inference Analysis}
Table~\ref{tab:sample_steps_performance} presents the inference latency and performance of our model across various sampling steps on the \emph{N24News} and \emph{MMHS11K} datasets. While performance improves slightly with more sampling steps, the inference time exhibits near-linear growth. Our model achieves effective manifold recovery at approximately 50 steps; beyond this threshold, increasing sampling steps yields diminishing returns in performance. Although our primary experiments utilize 50-step inference to ensure optimal quality, the results suggest a favorable trade-off for practical applications: 
reducing sampling steps to 20 or 30 can significantly accelerate generation with only marginal performance degradation. Figure~\ref{fig:inference_efficiency} compares the computational cost and inference latency of our model against baseline methods. 
The results indicate that while our method is slower than some competitors at 20 sampling steps, it maintains a low computational overhead. 
Consequently, optimizing the sampling rate and inference efficiency of diffusion-based models remains a critical direction for future investigation.

\section{Conclusion}

This paper proposes a restoration strategy that aims to addressing the challenges of missing modalities. We leverage an improved Diffusion model to effectively restore missing features, serving as a restorative mid-stage training module that significantly enhances the robustness and generalization of foundational Vision Language Models (VLMs) under modality-incomplete scenarios. The framework comprises two key components: a Dynamic Modality Gating mechanism and a Cross-modal Mutual Learning mechanism. Extensive experiments conducted across four benchmark datasets demonstrate that our model serves as an effective extension of VLMs for handling missing modalities across all scenarios.


\bibliographystyle{ACM-Reference-Format}
\bibliography{main}

\appendix

\end{document}